\documentclass[11pt,a4paper]{article}
\pdfoutput=1
\usepackage[nohyperref]{acl2018}
\usepackage{times}
\usepackage{latexsym}
\usepackage{amsmath}
\usepackage{amssymb}
\usepackage{booktabs}
\usepackage{url}
\usepackage{tikz}
\usepackage{tikz-dependency}
\usepackage{xspace}

\usetikzlibrary{graphs}
\usetikzlibrary{shapes}

\aclfinalcopy %
\def\aclpaperid{1268} %

\newcommand{\W}{\mathbf{W}}

\newcommand{\h}{\mathbf{h}}
\newcommand{\bb}{\mathbf{b}}
\newcommand{\x}{\mathbf{x}}
\newcommand{\z}{\mathbf{z}}
\newcommand{\rr}{\mathbf{r}}
\newcommand{\X}{\mathbf{X}}

\newcommand{\chrf}{{\sc chr}F++\xspace}
\newcommand{\stos}{\texttt{s2s}\xspace}
\newcommand{\gtos}{\texttt{g2s}\xspace}
\newcommand{\gtosplus}{\texttt{g2s+}\xspace}

\DeclareFontShape{OT1}{cmtt}{bx}{n}{<5><6><7><8><9><10><10.95><12><14.4><17.28><20.74><24.88>cmttb10}{}
\usepackage{color}

\title{Graph-to-Sequence Learning using Gated Graph Neural Networks}

\author{Daniel Beck$^\dagger$ ~~~~~ Gholamreza Haffari$^\ddagger$ ~~~~~ Trevor Cohn$^\dagger$\\
$^\dagger$School of Computing and Information Systems\\
University of Melbourne, Australia \\
{\tt \{d.beck,t.cohn\}@unimelb.edu.au} \\
$^\ddagger$Faculty of Information Technology \\
Monash University, Australia \\
{\tt gholamreza.haffari@monash.edu}\\}

\date{}

\begin{document}

\maketitle
\begin{abstract}

Many NLP applications can be framed as a graph-to-sequence learning problem. Previous work proposing neural architectures on this setting obtained promising results compared to grammar-based approaches but still rely on linearisation heuristics and/or standard recurrent networks to achieve the best performance. In this work, we propose a new model that encodes the full structural information contained in the graph. Our architecture couples the recently proposed Gated Graph Neural Networks with an input transformation that allows nodes and edges to have their own hidden representations, while tackling the parameter explosion problem present in previous work. Experimental results show that our model outperforms strong baselines in generation from AMR graphs and syntax-based neural machine translation.

\end{abstract}

\section{Introduction}

Graph structures are ubiquitous in representations of natural language. In particular, many whole-sentence semantic frameworks employ directed acyclic graphs as the underlying formalism, while most tree-based syntactic representations can also be seen as graphs. A range of NLP applications can be framed as the process of transducing a graph structure into a sequence. For instance, language generation may involve realising a semantic graph into a surface form and %
syntactic machine translation involves transforming a tree-annotated source sentence to its translation.

Previous work in this setting rely on grammar-based approaches such as tree transducers \cite{Flanigan2016} and hyperedge replacement grammars \cite{Jones2012}. A key limitation of these approaches is that alignments between graph nodes and surface tokens are required. These alignments are usually automatically generated so they can propagate errors when building the grammar. More recent approaches transform the graph into a linearised form and use off-the-shelf methods such as phrase-based machine translation \cite{Pourdamghani2016} or neural sequence-to-sequence (henceforth, \stos) models \cite{Konstas2017}. Such approaches ignore the full graph structure, discarding key information.

In this work we propose a model for graph-to-sequence (henceforth, \gtos) learning that leverages recent advances in neural encoder-decoder architectures. Specifically, we employ an encoder based on Gated Graph Neural Networks \cite[][GGNNs]{Li2016}, which can incorporate the full graph structure without loss of information. Such networks represent edge information as label-wise parameters, which can be problematic even for small sized label vocabularies (in the order of hundreds). To address this limitation, we also introduce a graph transformation that changes edges to additional nodes, solving the parameter explosion problem. This also ensures that edges have graph-specific hidden vectors, which gives more information to the attention and decoding modules in the network.

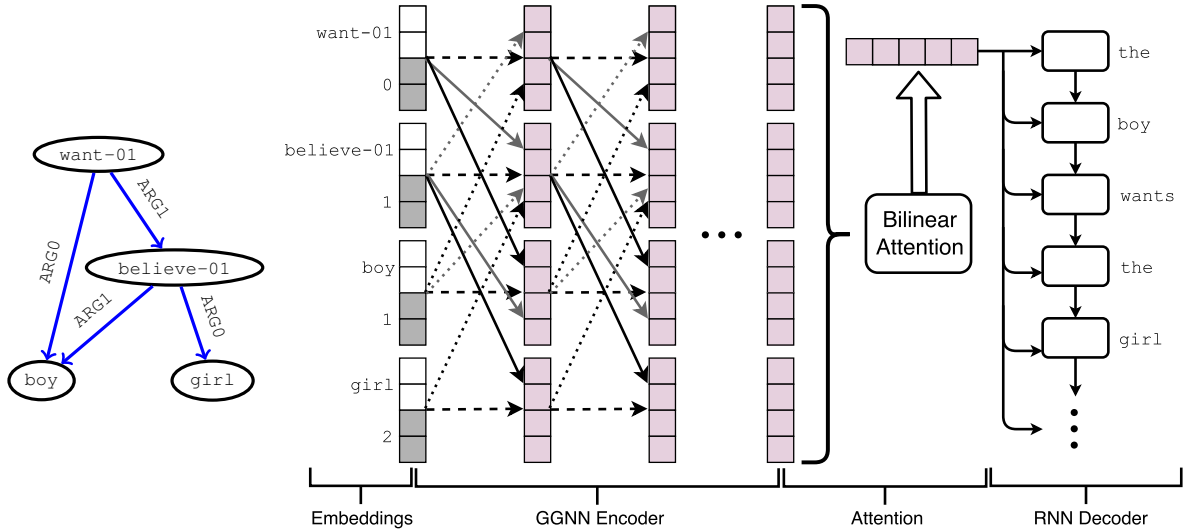
\begin{figure*}[t!]
  \centering
  \hspace{-0.3cm}\begin{minipage}{0.25\linewidth}
  {\tt \scriptsize
  \begin{tikzpicture}[very thick]
    \node[ellipse,draw=black] (want) at (0,0) {want-01};
    \node[ellipse,draw=black] (believe) at (1.0,-1.5) {believe-01};
    \node[ellipse,draw=black] (boy) at (-0.75,-3.0) {boy};
    \node[ellipse,draw=black] (girl) at (1.5,-3.0) {girl};

    \draw[->,draw=blue] (want) -> (boy) node[midway,sloped,above=0.1cm] {ARG0};
    \draw[->,draw=blue] (want) -> (believe) node[midway,sloped,above=0.1cm] {ARG1};
    \draw[->,draw=blue] (believe) -> (girl) node[midway,sloped,above=0.1cm] {ARG0};
    \draw[->,draw=blue] (believe) -> (boy) node[midway,sloped,above=0.1cm] {ARG1};

  \end{tikzpicture}
  }    
  \end{minipage}
  \hspace{-0.5cm}\begin{minipage}{0.7\linewidth}
  \includegraphics[scale=0.49]{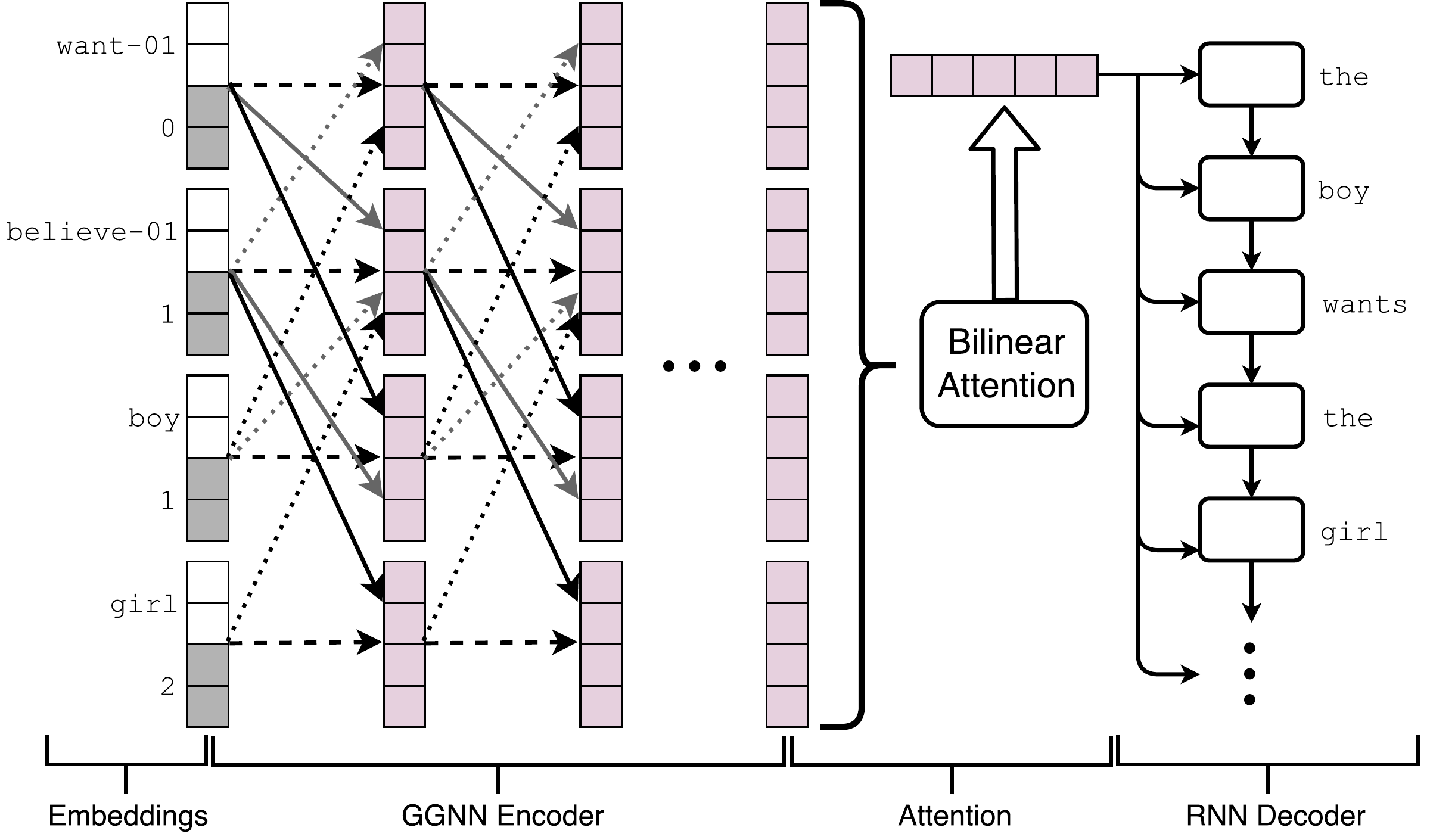}    
  \end{minipage}
  \caption{Left: the AMR graph representing the sentence ``The boy wants the girl to believe him.''. Right: Our proposed architecture using the same AMR graph as input and the surface form as output. The first layer is a concatenation of node and positional embeddings, using distance from the root node as the position. The GGNN encoder updates the embeddings using edge-wise parameters, represented by different colors (in this example, {\tt ARG0} and {\tt ARG1}). The encoder also add corresponding reverse edges (dotted arrows) and self edges for each node (dashed arrows). All parameters are shared between layers. Attention and decoder components are similar to standard \stos ~models. This is a pictorial representation: in our experiments the graphs are transformed before being used as inputs (see \S \ref{sec:bipartite}).}
  \label{fig:arch}
\end{figure*}

We benchmark our model in two graph-to-sequence problems, generation from Abstract Meaning Representations (AMRs) and Neural Machine Translation (NMT) with source dependency information. Our approach outperforms strong \stos ~baselines in both tasks {\em without relying on standard RNN encoders}, in contrast with previous work. In particular, for NMT we show that we avoid the need for RNNs by adding sequential edges between contiguous words in the dependency tree. This illustrates the generality of our approach: linguistic biases can be added to the inputs by simple graph transformations, without the need for changes to the model architecture.

\section{Neural Graph-to-Sequence Model}
\label{sec:g2s}

Our proposed architecture is shown in Figure \ref{fig:arch}, with an example AMR graph and its transformation into its surface form. Compared to standard \stos ~models, the main difference is in the encoder, where we employ a GGNN to build a graph representation. In the following we explain the components of this architecture in detail.\footnote{Our implementation uses MXNet \cite{Chen2015} and is based on the Sockeye toolkit \cite{Hieber2017}. Code is available at \url{github.com/beckdaniel/acl2018_graph2seq}.}

\subsection{Gated Graph Neural Networks}
\label{sec:ggnn}

Early approaches for recurrent networks on graphs \cite{Gori2005,Scarselli2009} assume a fixed point representation of the parameters and learn using contraction maps. \newcite{Li2016} argues that this restricts the capacity of the model and makes it harder to learn long distance relations between nodes. To tackle these issues, they propose Gated Graph Neural Networks, which extend these architectures with gating mechanisms in a similar fashion to Gated Recurrent Units \cite{Cho2014}. This allows the network to be learnt via modern backpropagation procedures.

In following, we formally define the version of GGNNs we employ in this study. Assume a directed graph $\mathcal{G} = \{ \mathcal{V},\mathcal{E}, L_{\mathcal{V}}, L_{\mathcal{E}} \}$, where $\mathcal{V}$ is a set of nodes $(v, \ell_v)$, $\mathcal{E}$ is a set of edges $(v_i, v_j, \ell_e )$ and $L_{\mathcal{V}}$ and $L_{\mathcal{E}}$ are respectively vocabularies for nodes and edges, from which node and edge labels ($\ell_v$ and $\ell_e$) are defined. %
Given an input graph with nodes mapped to embeddings $\X$, a GGNN is defined as
\begin{align*}
  \h_v^{0} &= \x_v \\
  \rr_v^t &= \sigma \left( c_v^r \sum\limits_{u \in \mathcal{N}_v} \W_{\ell_e}^r \h_u^{(t-1)} + \bb_{\ell_e}^r \right)\\
  \z_v^t &= \sigma \left( c_v^z \sum\limits_{u \in \mathcal{N}_v} \W_{\ell_e}^z \h_u^{(t-1)} + \bb_{\ell_e}^z \right)\\
  \widetilde\h_v^t &= \rho \left( c_v \sum\limits_{u \in \mathcal{N}_v} \W_{\ell_e} \left( \rr_u^t \odot \h_u^{(t-1)}\right) + \bb_{\ell_e} \right)\\
  \h_v^{t} &= (1 - \z_v^t) \odot \h_v^{(i-1)} + \z_v^t \odot \widetilde\h_v^t
\end{align*}
where $e = (u,v,\ell_e)$ is the edge between nodes $u$ and $v$, $\mathcal{N}(v)$ is the set of neighbour nodes for $v$, $\rho$ is a non-linear function, $\sigma$ is the sigmoid function and $c_v = c_v^z = c_v^r = |\mathcal{N}_v|^{-1}$ are normalisation constants.

Our formulation differs from the original GGNNs from \newcite{Li2016} in some aspects: 1) we add bias vectors for the hidden state, reset gate and update gate computations; 2) label-specific matrices do not share any components; 3) reset gates are applied to all hidden states before any computation and 4) we add normalisation constants. These modifications were applied based on preliminary experiments and ease of implementation.

An alternative to GGNNs is the model from \newcite{Marcheggiani2017}, which add edge label information to Graph Convolutional Networks (GCNs).
According to \newcite{Li2016}, the main difference between GCNs and GGNNs is analogous to the difference between convolutional and recurrent networks. More specifically, GGNNs can be seen as multi-layered GCNs where layer-wise parameters are tied and gating mechanisms are added. A large number of layers can propagate node information between longer distances in the graph and, unlike GCNs, GGNNs can have an arbitrary number of layers without increasing the number of parameters. Nevertheless, our architecture borrows ideas from GCNs as well, such as normalising factors.

\subsection{Using GGNNs in attentional encoder-decoder models}
\label{sec:ggrn-based-encoder}

In \stos ~models, inputs are sequences of tokens where each token is represented by an embedding vector. The encoder then transforms these vectors into hidden states by incorporating context, usually through a recurrent or a convolutional network. These are fed into an attention mechanism, generating a single context vector that informs decisions in the decoder.

Our model follows a similar structure, where the encoder is a GGNN that receives {\em node} embeddings as inputs and generates {\em node} hidden states as outputs, using the graph structure as context. This is shown in the example of Figure \ref{fig:arch}, where we have 4 hidden vectors, one per node in the AMR graph. The attention and decoder components follow similar standard \stos ~models, where we use a bilinear attention mechanism \cite{Luong2015} and a 2-layered LSTM \cite{Hochreiter1997} as the decoder. Note, however, that other decoders and attention mechanisms can be easily employed instead. \newcite{Bastings2017} employs a similar idea for syntax-based NMT, but using GCNs instead.

\subsection{Bidirectionality and positional embeddings}
\label{sec:posit-embed}

While our architecture can in theory be used with general graphs, rooted directed acyclic graphs (DAGs) are arguably the most common kind in the problems we are addressing. This means that node embedding information is propagated in a top down manner. However, it is desirable to have information flow from the reverse direction as well, in the same way RNN-based encoders benefit from right-to-left propagation (as in bidirectional RNNs). \newcite{Marcheggiani2017} and \newcite{Bastings2017} achieve this by adding reverse edges to the graph, as well as self-loops edges for each node. These extra edges have specific labels, hence their own parameters in the network.

In this work, we also follow this procedure to ensure information is evenly propagated in the graph. However, this raises another limitation: because the graph becomes essentially undirected, the encoder is now unaware of any intrinsic hierarchy present in the input. %
Inspired by \newcite{Gehring2017} and \newcite{Vaswani2017}, we tackle this problem by adding positional embeddings to every node. These embeddings are indexed by integer values representing the minimum distance from the root node and are learned as model parameters.\footnote{\newcite{Vaswani2017} also proposed fixed positional embeddings based on sine and cosine wavelengths. Preliminary experiments showed that this approach did not work in our case: we speculate this is because wavelengths are more suitable to sequential inputs.}
This kind of positional embedding is restricted to rooted DAGs: for general graphs, different notions of distance could be employed.

\section{Levi Graph Transformation}
\label{sec:bipartite}

\begin{figure}[t!]
  \centering
  \begin{scriptsize}
    {\tt
  \begin{tikzpicture}[very thick]
    \node[ellipse,draw=black] (want) at (0,0) {want-01};
    \node[ellipse,draw=black] (believe) at (0,-1.25) {believe-01};
    \node[ellipse,draw=black] (boy) at (0,-2.5) {boy};
    \node[ellipse,draw=black] (girl) at (0,-3.75) {girl};

    \node[ellipse,draw=black] (arg10) at (4,0) {ARG1};
    \node[ellipse,draw=black] (arg00) at (4,-1.25) {ARG0};
    \node[ellipse,draw=black] (arg11) at (4,-2.5) {ARG1};
    \node[ellipse,draw=black] (arg01) at (4,-3.75) {ARG0};

    \draw[->,draw=blue] (want) -> (arg00);
    \draw[->,draw=blue] (want) -> (arg10);
    \draw[->,draw=blue] (believe) -> (arg01);
    \draw[->,draw=blue] (believe) -> (arg11);
    \draw[->,draw=blue] (arg00) -> (boy);
    \draw[->,draw=blue] (arg10) -> (believe);
    \draw[->,draw=blue] (arg01) -> (girl);
    \draw[->,draw=blue] (arg11) -> (boy);

  \end{tikzpicture}
  \vspace{0.5cm}
  \hrule
  \vspace{0.5cm}
  \begin{tikzpicture}[very thick]
    \node[ellipse,draw=black] (want) at (0,0) {want-01};
    \node[ellipse,draw=black] (believe) at (0,-1.5) {believe-01};
    \node[ellipse,draw=black] (boy) at (0,-3) {boy};
    \node[ellipse,draw=black] (girl) at (0,-4.5) {girl};

    \node[ellipse,draw=black] (arg10) at (4,0) {ARG1};
    \node[ellipse,draw=black] (arg00) at (4,-1.5) {ARG0};
    \node[ellipse,draw=black] (arg11) at (4,-3) {ARG1};
    \node[ellipse,draw=black] (arg01) at (4,-4.5) {ARG0};

    \draw[->,draw=blue] (want) to [bend left=10](arg00);
    \draw[->,draw=blue] (want) to [bend left=10](arg10);
    \draw[->,draw=blue] (believe) to [bend left=10](arg01);
    \draw[->,draw=blue] (believe) to [bend left=10](arg11);
    \draw[->,draw=blue] (arg00) to [bend left=10](boy);
    \draw[->,draw=blue] (arg10) to [bend left=10](believe);
    \draw[->,draw=blue] (arg01) to [bend left=10](girl);
    \draw[->,draw=blue] (arg11) to [bend left=10](boy);

    \draw[dashed,->,draw=red] (arg00) to [bend left=10](want);
    \draw[dashed,->,draw=red] (arg10) to [bend left=10](want);
    \draw[dashed,->,draw=red] (arg01) to [bend left=10](believe);
    \draw[dashed,->,draw=red] (arg11) to [bend left=10](believe);
    \draw[dashed,->,draw=red] (boy) to [bend left=10](arg00);
    \draw[dashed,->,draw=red] (believe) to [bend left=10](arg10);
    \draw[dashed,->,draw=red] (girl) to [bend left=10](arg01);
    \draw[dashed,->,draw=red] (boy) to [bend left=10](arg11);

    \draw[dotted,->,draw=purple] (want) to [loop left](want);
    \draw[dotted,->,draw=purple] (believe) to [loop left](believe);
    \draw[dotted,->,draw=purple] (boy) to [loop left](boy);
    \draw[dotted,->,draw=purple] (girl) to [loop left](girl);
    \draw[dotted,->,draw=purple] (arg00) to [loop right](arg00);
    \draw[dotted,->,draw=purple] (arg01) to [loop right](arg01);
    \draw[dotted,->,draw=purple] (arg10) to [loop right](arg10);
    \draw[dotted,->,draw=purple] (arg11) to [loop right](arg11);

  \end{tikzpicture}
  }
  \end{scriptsize}
  \caption{Top: the AMR graph from Figure \ref{fig:arch} transformed into its corresponding Levi graph. Bottom: Levi graph with added reverse and self edges (colors represent different edge labels).}
  \label{fig:bip}
\end{figure}

The \gtos ~model proposed in \S \ref{sec:g2s} has two key deficiencies. First, GGNNs have three linear transformations {\em per edge type}. This means that the number of parameters can explode: AMR, for instance, has around 100 different predicates, which correspond to edge labels. Previous work deal with this problem by explicitly grouping edge labels into a single one \cite{Marcheggiani2017,Bastings2017} but this is not an ideal solution since it incurs in loss of information.

The second deficiency is that edge label information is encoded in the form of GGNN parameters in the network. This means that each label will have the same ``representation'' across all graphs. However, the latent information in edges can depend on the content in which they appear in a graph. %
Ideally, edges should have instance-specific {\em hidden states}, in the same way as nodes, and these should also inform decisions made in the decoder through the attention module. For instance, in the AMR graph shown in Figure \ref{fig:arch}, the {\tt ARG1} predicate between {\tt want-01} and {\tt believe-01} can be interpreted as the preposition ``to'' in the surface form, while the {\tt ARG1} predicate connecting {\tt believe-01} and {\tt boy} is realised as a pronoun. Notice that edge hidden vectors are already present in \stos ~networks that use linearised graphs: we would like our architecture to also have this benefit.

Instead of modifying the architecture, we propose to transform the input graph into its equivalent Levi graph \cite[][p. 765]{Levi1942,Gross2004}. Given a graph $\mathcal{G} = \{ \mathcal{V},\mathcal{E}, L_{\mathcal{V}}, L_{\mathcal{E}} \}$, a Levi graph\footnote{Formally, a Levi graph is defined over any {\em incidence structure}, which is a general concept usually considered in a geometrical context. Graphs are an example of incidence structures but so are points and lines in the Euclidean space, for instance.} is defined as $\mathcal{G} = \{ \mathcal{V}', \mathcal{E}', L_{\mathcal{V}'}, L_{\mathcal{E}'} \}$, where $\mathcal{V}' = \mathcal{V} \cup \mathcal{E}$, $L_{\mathcal{V}'} = L_{\mathcal{V}} \cup L_{\mathcal{E}}$ and $L_{\mathcal{E}'} = \varnothing$. The new edge set $\mathcal{E}'$ contains a edge for every (node, edge) pair that is present in the original graph. By definition, the Levi graph is bipartite.

Intuitively, transforming a graph into its Levi graph equivalent turns edges into additional nodes. While simple in theory, this transformation addresses both modelling deficiencies mentioned above in an elegant way. Since the Levi graph has no labelled edges there is no risk of parameter explosion: original edge labels are represented as embeddings, in the same way as nodes. Furthermore, the encoder now naturally generates hidden states for original edges as well.

In practice, we follow the procedure in \S \ref{sec:posit-embed} and add reverse and self-loop edges to the Levi graph, so the practical edge label vocabulary is $L_{\mathcal{E}'} = \{\mathrm{default,reverse,self}\}$. This still keeps the parameter space modest since we have only three labels. Figure \ref{fig:bip} shows the transformation steps in detail, applied to the AMR graph shown in Figure \ref{fig:arch}. Notice that the transformed graphs are the ones fed into our architecture: we show the original graph in Figure \ref{fig:arch} for simplicity.

It is important to note that this transformation can be applied to any graph and therefore is independent of the model architecture. We speculate this can be beneficial in other kinds of graph-based encoder such as GCNs and leave further investigation to future work.

\section{Generation from AMR Graphs}
\label{sec:amr}

Our first \gtos ~benchmark is language generation from AMR, a semantic formalism that represents sentences as rooted DAGs \cite{Banarescu2013}. Because AMR abstracts away from syntax, graphs do not have gold-standard alignment information, so generation is not a trivial task. Therefore, we hypothesize that our proposed model is ideal for this problem.

\subsection{Experimental setup}
\label{sec:amrexp}

\paragraph{Data and preprocessing}
\label{sec:amrdata}

We use the latest AMR corpus release (LDC2017T10) with the default split of 36521/1368/1371 instances for training, development and test sets. Each graph is preprocessed using a procedure similar to what is performed by \newcite{Konstas2017}, which includes entity simplification and anonymisation. This preprocessing is done before transforming the graph into its Levi graph equivalent. For the \stos ~baselines, we also add scope markers as in \newcite{Konstas2017}. We detail these procedures in the Appendix.

\paragraph{Models}
\label{sec:amrmodels}

Our baselines are attentional \stos ~models which take linearised graphs as inputs. The architecture is similar to the one used in \newcite{Konstas2017} for AMR generation, where the encoder is a BiLSTM followed by a unidirectional LSTM. All dimensionalities are fixed to 512.

For the \gtos ~models, %
we fix the number of layers in the GGNN encoder to 8, as this gave the best results on the development set. Dimensionalities are also fixed at 512 except for the GGNN encoder which uses 576. This is to ensure all models have a comparable number of parameters and therefore similar capacity.

Training for all models uses Adam \cite{Kingma2015} with 0.0003 initial learning rate and 16 as the batch size.\footnote{Larger batch sizes hurt dev performance in our preliminary experiments. There is evidence that small batches can lead to better generalisation performance \cite{Keskar2017}. While this can make training time slower, it was doable in our case since the dataset is small.} To regularise our models we perform early stopping on the dev set based on perplexity and apply 0.5 dropout \cite{Srivastava2014} on the source embeddings. We detail additional model and training hyperparameters in the Appendix.

\paragraph{Evaluation}
\label{sec:eval}

Following previous work, we evaluate our models using BLEU \cite{Papineni2001} and perform bootstrap resampling to check statistical significance. However, since recent work has questioned the effectiveness of BLEU with bootstrap resampling \cite{Graham2014}, we also report results using sentence-level \chrf ~\cite{Popovic2017}, using the Wilcoxon signed-rank test to check significance. Evaluation is case-insensitive for both metrics. %

Recent work has shown that evaluation in neural models can lead to wrong conclusions by just changing the random seed \cite{Reimers2017}. In an effort to make our conclusions more robust, we run each model 5 times using different seeds. %
From each pool, we report results using the median model according to performance on the dev set (simulating what is expected from a single run) and using an ensemble of the 5 models. %

Finally, we also report the number of parameters used in each model. Since our encoder architectures are quite different, we try to match the number of parameters between them by changing the dimensionality of the hidden layers (as explained above). We do this to minimise the effects of model capacity as a confounder.

\subsection{Results and analysis}
\label{sec:amrres}

Table \ref{tab:amr} shows the results on the test set. For the \stos ~models, we also report results without the scope marking procedure of \newcite{Konstas2017}. %
Our approach significantly outperforms the \stos ~baselines both with individual models and ensembles, while using a comparable number of parameters. In particular, we obtain these results without relying on scoping heuristics.

\begin{table}[t!]
  \centering
  \begin{tabular}{lccc}
    \toprule
    & BLEU & \chrf & \#params \\
    \midrule
    \multicolumn{4}{l}{\em Single models} \\
    \stos & 21.7 & 49.1 & 28.4M  \\
    \stos (-s) & 18.4 & 46.3 & 28.4M \\
    \gtos & 23.3 & 50.4 & 28.3M\\
    \midrule
    \multicolumn{4}{l}{\em Ensembles} \\
    \stos & 26.6 & 52.5 & 142M \\
    \stos (-s) & 22.0 & 48.9 & 142M \\
    \gtos & \bf 27.5 & \bf 53.5 & 141M \\
    \midrule
    \multicolumn{4}{l}{\em Previous work (early AMR treebank versions)} \\
    KIYCZ17 & 22.0 & -- & -- \\
    \multicolumn{4}{l}{\em Previous work (as above + unlabelled data)} \\
    KIYCZ17 & 33.8 & -- & -- \\
    PKH16 & 26.9 & -- & -- \\
    SPZWG17 & 25.6 & -- & -- \\
    FDSC16 & 22.0 & -- & --\\
    \bottomrule
  \end{tabular}
  \caption{Results for AMR generation on the test set. All score differences between our models and the  corresponding baselines are significantly different (p$<$0.05). %
``(-s)'' means input without scope marking. KIYCZ17, PKH16, SPZWG17 and FDSC16 are respectively the results reported in \newcite{Konstas2017}, \newcite{Pourdamghani2016}, \newcite{Song2017} and \newcite{Flanigan2016}.}
  \label{tab:amr}
\end{table}

On Figure \ref{fig:example} we show an example where our model outperforms the baseline. The AMR graph contains four reentrancies, predicates that reference previously defined concepts in the graph. In the \stos models including \newcite{Konstas2017}, reentrant nodes are copied in the linearised form, while this is not necessary for our \gtos models. We can see that the \stos prediction overgenerates the ``India and China'' phrase. The \gtos prediction avoids overgeneration, and almost perfectly matches the reference. While this is only a single example, it provides evidence that retaining the full graphical structure is beneficial for this task, which is corroborated by our quantitative results.

\begin{figure}[t!]
  \centering
  \begin{tabular}{p{7.3cm}}
\toprule
    Original AMR graph \\
    \small \tt (p / propose-01 
    
    ~ :ARG0 (c / country 
    
    ~~ :wiki "Russia" 

    ~~ :name (n / name %

    ~~~ :op1 "Russia"))

    ~ :ARG1 (c5 / cooperate-01 

    ~~ :ARG0 c

    ~~ :ARG1 (a / and 
    
    ~~~ :op1 \textbf{(c2 / country}
    
    ~~~~ \textbf{:wiki "India"}

    ~~~~ \textbf{:name (n2 / name} %

    ~~~~~ \textbf{:op1 "India"))}
    
    ~~~ :op2 \textbf{(c3 / country}

    ~~~~ \textbf{:wiki "China"}

    ~~~~ \textbf{:name (n3 / name} %

    ~~~~~ \textbf{:op1 "China"))}))

    ~ :purpose (i / increase-01 

    ~~ :ARG0 c5 

    ~~ :ARG1 (s / security) 

    ~~ :location (a2 / around 
    
    ~~~ :op1 (c4 / country 

    ~~~~ :wiki "Afghanistan" 

    ~~~~ :name (n4 / name 

    ~~~~~ :op1 "Afghanistan")))

    ~~ :purpose (b / block-01 

    ~~~ :ARG0 (a3 / and 
    
    ~~~~ :op1 c \textbf{:op2 c2 :op3 c3}

    ~~~ :ARG1 (s2 / supply-01 

    ~~~~ :ARG1 (d / drug))))) \\
    \midrule
    Reference surface form \\
    \small Russia proposes cooperation with {\bf India and China} to increase security around Afghanistan to block drug supplies. \\
    \midrule
    \stos output (\chrf 61.8)\\
    \small Russia proposed cooperation with {\bf India and China} to increase security around the Afghanistan to block security around the Afghanistan , {\bf India and China}. \\
    \midrule 
    \gtos output (\chrf 78.2)\\
    \small Russia proposed cooperation with {\bf India and China} to increase security around Afghanistan to block drug supplies. \\
\bottomrule
  \end{tabular}
  \caption{Example showing overgeneration due to reentrancies. Top: original AMR graph with key reentrancies highlighted. Bottom: reference and outputs generated by the \stos and \gtos models, highlighting the overgeneration phenomena.}
  \label{fig:example}
\end{figure}

Table \ref{tab:amr} also show BLEU scores reported in previous work. These results are not strictly comparable because they used different training set versions and/or employ additional unlabelled corpora; nonetheless some insights can be made. In particular, our \gtos ~ensemble performs better than many previous models that combine a smaller training set with a large unlabelled corpus. It is also most informative to compare our \stos ~model with \newcite{Konstas2017}, since this baseline is very similar to theirs. We expected our single model baseline to outperform theirs since we use a larger training set but we obtained similar performance. We speculate that better results could be obtained by more careful tuning, but nevertheless we believe such tuning would also benefit our proposed \gtos ~architecture.

The best results with unlabelled data are obtained by \newcite{Konstas2017} using Gigaword sentences as additional data and a paired trained procedure with an AMR parser. It is important to note that this procedure is orthogonal to the individual models used for generation and parsing. Therefore, we hypothesise that our model can also benefit from such techniques, an avenue that we leave for future work.

\section{Syntax-based Neural Machine Translation}
\label{sec:syntax-based-nmt}

Our second evaluation is NMT, using as graphs source language dependency syntax trees. We focus on a medium resource scenario where additional linguistic information tends to be more beneficial. Our experiments comprise two language pairs: English-German and English-Czech.

\subsection{Experimental setup}
\label{sec:nmtexp}

\paragraph{Data and preprocessing}

We employ the same data and settings from \newcite{Bastings2017},\footnote{We obtained the data from the original authors to ensure results are comparable without any influence from preprocessing steps.} which use the News Commentary V11 corpora from the WMT16 translation task.\footnote{\url{http://www.statmt.org/wmt16/translation-task.html}} English text is tokenised and parsed using SyntaxNet\footnote{\url{https://github.com/tensorflow/models/tree/master/syntaxnet}} while German and Czech texts are tokenised and split into subwords using byte-pair encodings \cite[][BPE]{Sennrich2016} (8000 merge operations). We refer to \newcite{Bastings2017} for further information on the preprocessing steps.

Labelled dependency trees in the source side are transformed into Levi graphs as a preprocessing step. However, unlike AMR generation, in NMT the inputs are originally surface forms that contain important sequential information. This information is lost when treating the input as dependency trees, which might explain why \newcite{Bastings2017} obtain the best performance when using an initial RNN layer in their encoder. To investigate this phenomenon, we also perform experiments adding sequential connections to each word in the dependency tree, corresponding to their order in the original surface form (henceforth, \gtosplus). These connections are represented as edges with specific $\mathrm{left}$ and $\mathrm{right}$ labels, which are added after the Levi graph transformation. Figure \ref{fig:nmt} shows an example of an input graph for \gtosplus, with the additional sequential edges connecting the words (reverse and self edges are omitted for simplicity).

\begin{figure}[ht!]
  \centering
  {\tt
  \begin{dependency}
    \begin{deptext}
      There \& is \& a \& deeper \& issue \& at \& stake \& . \\
    \end{deptext}
    \deproot[edge unit distance=4ex]{2}{ROOT}
    \depedge{2}{1}{expl}
    \depedge{2}{5}{nsubj}
    \depedge[edge unit distance=2ex]{2}{8}{punct}
    \depedge{5}{3}{det}
    \depedge{5}{4}{amod}
    \depedge{5}{6}{prep}
    \depedge{6}{7}{pobj}
  \end{dependency}
  }

  {\tt \footnotesize
  \begin{tikzpicture}[very thick]
    \node[ellipse,draw=black] (there) at (0,0) {There};
    \node[ellipse,draw=black] (is) at (0,-1.25) {is};
    \node[ellipse,draw=black] (a) at (0,-2.5) {a};
    \node[ellipse,draw=black] (deeper) at (0,-3.75) {deeper};
    \node[ellipse,draw=black] (issue) at (0,-5) {issue};
    \node[ellipse,draw=black] (at) at (0,-6.25) {at};
    \node[ellipse,draw=black] (stake) at (0,-7.5) {stake};
    \node[ellipse,draw=black] (dot) at (0,-8.75) {.};

    \node[ellipse,draw=black] (ROOT) at (4,0) {ROOT};
    \node[ellipse,draw=black] (expl) at (4,-1.25) {expl};
    \node[ellipse,draw=black] (nsubj) at (4,-2.5) {nsubj};
    \node[ellipse,draw=black] (punct) at (4,-3.75) {punct};
    \node[ellipse,draw=black] (det) at (4,-5) {det};
    \node[ellipse,draw=black] (amod) at (4,-6.25) {amod};
    \node[ellipse,draw=black] (prep) at (4,-7.5) {prep};
    \node[ellipse,draw=black] (pobj) at (4,-8.75) {pobj};

    \draw[->,draw=blue] (ROOT) -> (is);
    \draw[->,draw=blue] (expl) -> (there);
    \draw[->,draw=blue] (nsubj) -> (issue);
    \draw[->,draw=blue] (punct) -> (dot);
    \draw[->,draw=blue] (det) -> (a);
    \draw[->,draw=blue] (amod) -> (deeper);
    \draw[->,draw=blue] (prep) -> (at);
    \draw[->,draw=blue] (pobj) -> (stake);

    \draw[->,draw=blue] (is) -> (expl);
    \draw[->,draw=blue] (is) -> (nsubj);
    \draw[->,draw=blue] (is) -> (punct);
    \draw[->,draw=blue] (issue) -> (det);
    \draw[->,draw=blue] (issue) -> (amod);
    \draw[->,draw=blue] (issue) -> (prep);
    \draw[->,draw=blue] (at) -> (pobj);

    \draw[dashed,->,draw=brown] (there) -> (is);
    \draw[dashed,->,draw=brown] (is) -> (a);
    \draw[dashed,->,draw=brown] (a) -> (deeper);
    \draw[dashed,->,draw=brown] (deeper) -> (issue);
    \draw[dashed,->,draw=brown] (issue) -> (at);
    \draw[dashed,->,draw=brown] (at) -> (stake);
    \draw[dashed,->,draw=brown] (stake) -> (dot);

  \end{tikzpicture}
  }
  \caption{Top: a sentence with its corresponding dependency tree. Bottom: the transformed tree into a Levi graph with additional sequential connections between words (dashed lines). The full graph also contains reverse and self edges, which are omitted in the figure.}
  \label{fig:nmt}
\end{figure}

\paragraph{Models}

Our \stos ~and \gtos ~models are almost  the same as in the AMR generation experiments (\S \ref{sec:amrexp}). The only exception is the GGNN encoder dimensionality, where we use 512 for the experiments with dependency trees only and 448 when the inputs have additional sequential connections. As in the AMR generation setting, we do this to ensure model capacity are comparable in the number of parameters. Another key difference is that the \stos ~baselines do not use dependency trees: they are trained on the sentences only.

In addition to neural models, we also report results for Phrase-Based Statistical MT (PB-SMT), using Moses \cite{Koehn2007}. The PB-SMT models are trained using the same data conditions as \stos (no dependency trees) and use the standard setup in Moses, except for the language model, where we use a 5-gram LM trained on the target side of the respective parallel corpus.\footnote{Note that target data is segmented using BPE, which is not the usual setting for PB-SMT. We decided to keep the segmentation to ensure data conditions are the same.}

\paragraph{Evaluation}

We report results in terms of BLEU and \chrf, using case-sensitive versions of both metrics. Other settings are kept the same as in the AMR generation experiments (\S \ref{sec:amrexp}). For PB-SMT, we also report the median result of 5 runs, obtained by tuning the model using MERT \cite{Och2002a} 5 times.

\subsection{Results and analysis}
\label{sec:nmtres}

Table \ref{tab:nmt} shows the results on the respective test set for both language pairs. The \gtos ~models, which do not account for sequential information, lag behind our baselines. This is in line with the findings of \newcite{Bastings2017}, who found that having a BiRNN layer was key to obtain the best results. However, the \gtosplus ~models outperform the baselines in terms of BLEU scores under the same parameter budget, in both single model and ensemble scenarios. This result show that it is possible to incorporate sequential biases in our model without relying on RNNs or any other modification in the architecture.

\begin{table}[t!]
  \centering
  \begin{tabular}{lccc}
    \bottomrule
    \multicolumn{4}{c}{\bf English-German} \\
    \toprule
    & BLEU & \chrf & \#params \\
    \midrule
    \multicolumn{4}{l}{{\em Single models}} \\
    PB-SMT & 12.8 & 43.2 & -- \\
    \stos & 15.5 & 40.8 & 41.4M  \\
    \gtos & 15.2 & 41.4 & 40.8M\\
    \gtosplus & 16.7 & 42.4 & 41.2M\\
    \midrule
    \multicolumn{4}{l}{{\em Ensembles}} \\
    \stos & 19.0 & 44.1 & 207M \\
    \gtos & 17.7 & 43.5 & 204M \\
    \gtosplus & \bf 19.6 & \bf 45.1 & 206M\\
    \midrule
    \multicolumn{4}{l}{{\em Results from} \cite{Bastings2017}} \\
    BoW+GCN & 12.2 & -- & -- \\
    BiRNN & 14.9 & -- & -- \\
    BiRNN+GCN & 16.1 & -- & -- \\
    \bottomrule
    \multicolumn{4}{c}{\bf English-Czech} \\
    \toprule
    & BLEU & \chrf & \#params \\
    \midrule
    \multicolumn{4}{l}{{\em Single models}} \\
    PB-SMT & 8.6 & \bf 36.4 & -- \\
    \stos & 8.9 & 33.8 & 39.1M  \\
    \gtos & 8.7 & 32.3 & 38.4M\\
    \gtosplus & 9.8 & 33.3 & 38.8M\\
    \midrule
    \multicolumn{4}{l}{{\em Ensembles}} \\
    \stos & 11.3 & \bf 36.4 & 195M \\
    \gtos & 10.4 & 34.7 & 192M \\
    \gtosplus & \bf 11.7 & 35.9 & 194M\\
    \midrule
    \multicolumn{4}{l}{{\em Results from} \cite{Bastings2017}} \\
    BoW+GCN & 7.5 & -- & -- \\
    BiRNN & 8.9 & -- & -- \\
    BiRNN+GCN & 9.6 & -- & -- \\
    \bottomrule
  \end{tabular}
  \caption{Results for syntax-based NMT on the test sets. All score differences between our models and the  corresponding baselines are significantly different (p$<$0.05), including the negative \chrf result for En-Cs.}%
  \label{tab:nmt}
\end{table}

\newpage

Interestingly, we found different trends when analysing the \chrf numbers. In particular, this metric favours the PB-SMT models for both language pairs, while also showing improved performance for \stos in En-Cs. \chrf has been shown to better correlate with human judgments compared to BLEU, both at system and sentence level for both language pairs \cite{Bojar2017}, which motivated our choice as an additional metric. We leave further investigation of this phenomena for future work.

We also show some of the results reported by \newcite{Bastings2017} in Table \ref{tab:nmt}. Note that their results were based on a different implementation, %
which may explain some variation in performance. Their BoW+GCN model is the most similar to ours, as it uses only an embedding layer and a GCN encoder. We can see that even our simpler \gtos model outperforms their results. A key difference between their approach and ours is the Levi graph transformation and the resulting hidden vectors for edges. We believe their architecture would also benefit from our proposed transformation. In terms of baselines, \stos ~performs better than their BiRNN model for En-De and comparably for En-Cs, which corroborates that our baselines are strong ones. Finally, our \gtosplus ~single models outperform their BiRNN+GCN results, in particular for En-De, which is further evidence that RNNs are not necessary for obtaining the best performance in this setting.

An important point about these experiments is that we did not tune the architecture: we simply employed the same model we used in the AMR generation experiments, only adjusting the dimensionality of the encoder to match the capacity of the baselines.
We speculate that even better results would be obtained by tuning the architecture to this task. Nevertheless, we still obtained improved performance over our baselines and previous work, underlining the generality of our architecture.

\section{Related work}
\label{sec:related-work}

\paragraph{Graph-to-sequence modelling} Early NLP approaches for this problem were based on Hyperedge Replacement Grammars \cite[][HRGs]{Drewes1997}. These grammars assume the transduction problem can be split into rules that map portions of a graph to a set of tokens in the output sequence. In particular, \newcite{Chiang2013} defines a parsing algorithm, followed by a complexity analysis, %
while \newcite{Jones2012} report experiments on semantic-based machine translation using HRGs. %
HRGs were also used in previous work on AMR parsing \cite{Peng2015}. The main drawback of these grammar-based approaches though is the need for alignments between graph nodes and surface tokens, which are usually not available in gold-standard form.

\paragraph{Neural networks for graphs} Recurrent networks on general graphs were first proposed under the name Graph Neural Networks \cite{Gori2005,Scarselli2009}. %
Our work is based on the architecture proposed by \newcite{Li2016}, which add gating mechanisms. The main difference between their work and ours is that they focus on problems that concern the input graph itself such as node classification or path finding while we focus on generating strings. The main alternative for neural-based graph representations is Graph Convolutional Networks \cite{Bruna2014,Duvenaud2015,Kipf2017}, which have been applied in a range of problems. In NLP, \newcite{Marcheggiani2017} use a similar architecture for Semantic Role Labelling. They use heuristics to mitigate the parameter explosion by grouping edge labels, while we keep the original labels through our Levi graph transformation. An interesting alternative is proposed by \newcite{Schlichtkrull2017}, which uses tensor factorisation to reduce the number of parameters.

\paragraph{Applications} Early work on AMR generation employs grammars and transducers \cite{Flanigan2016,Song2017}. %
Linearisation approaches include \cite{Pourdamghani2016} and \cite{Konstas2017}, which showed that graph simplification and anonymisation are key to good performance, a procedure we also employ in our work. However, compared to our approach, linearisation incurs in loss of information. MT has a long history of previous work that aims at incorporating syntax  \cite[][inter alia]{Wu1997,Yamada2001,Galley2004,Liu2006}. This idea has also been investigated in the context of NMT. \newcite{Bastings2017} is the most similar work to ours, and we benchmark against their approach in our NMT experiments. \newcite{Eriguchi2016} also employs source syntax, but using constituency trees instead. Other approaches have investigated the use of syntax in the target language \cite{Aharoni2017,Eriguchi2017}. Finally, \newcite{Hashimoto2017} treats source syntax as a latent variable, which can be pretrained using annotated data.

\section{Discussion and Conclusion}
\label{sec:conclusion}

We proposed a novel encoder-decoder architecture for graph-to-sequence learning, outperforming baselines in two NLP tasks: generation from AMR graphs and syntax-based NMT. Our approach addresses shortcomings from previous work, including loss of information from linearisation and parameter explosion. In particular, we showed how graph transformations can solve issues with graph-based networks without changing the underlying architecture. This is the case of the proposed Levi graph transformation, which ensures the decoder can attend to edges as well as nodes, but also to the sequential connections added to the dependency trees in the case of NMT. Overall, because our architecture can work with general graphs, it is straightforward to add linguistic biases in the form of extra node and/or edge information. We believe this is an interesting research direction in terms of applications.

Our architecture nevertheless has two major limitations. %
The first one is that GGNNs have a fixed number of layers, even though graphs can vary in size in terms of number of nodes and edges. %
A better approach would be to allow the encoder to have a dynamic number of layers, possibly based on the diameter (longest path) in the input graph. %
The second limitation comes from the Levi graph transformation: because edge labels are represented as nodes they end up sharing the vocabulary and therefore, the same semantic space. This is not ideal, as nodes and edges are different entities. %
An interesting alternative is Weave Module Networks \cite{Kearnes2016}, which explicitly decouples node and edge representations without incurring in parameter explosion. Incorporating both ideas to our architecture is an %
research direction we plan for future work.

\section*{Acknowledgements}

This work was supported by the Australian Research Council (DP160102686). 
The research reported in this paper was partly conducted at the 2017 Frederick Jelinek Memorial Summer Workshop on Speech and Language Technologies, hosted at Carnegie Mellon University and sponsored by Johns Hopkins University with unrestricted gifts from Amazon, Apple, Facebook, Google, and Microsoft. The authors would also like to thank Joost Bastings for sharing the data from his paper's experiments.

\bibliography{nn,nmt,sem,tools,syntax,books,genmt}

\begin{thebibliography}{}
\expandafter\ifx\csname natexlab\endcsname\relax\def\natexlab#1{#1}\fi

\bibitem[{Aharoni and Goldberg(2017)}]{Aharoni2017}
Roee Aharoni and Yoav Goldberg. 2017.
\newblock {Towards String-to-Tree Neural Machine Translation}.
\newblock In {\em Proceedings of ACL\/}. pages 132--140.

\bibitem[{Banarescu et~al.(2013)Banarescu, Bonial, Cai, Georgescu, Griffitt,
  Hermjakob, Knight, Koehn, Palmer, and Schneider}]{Banarescu2013}
Laura Banarescu, Claire Bonial, Shu Cai, Madalina Georgescu, Kira Griffitt, Ulf
  Hermjakob, Kevin Knight, Philipp Koehn, Martha Palmer, and Nathan Schneider.
  2013.
\newblock {Abstract Meaning Representation for Sembanking}.
\newblock In {\em Proceedings of the 7th Linguistic Annotation Workshop and
  Interoperability with Discourse\/}. pages 178--186.

\bibitem[{Bastings et~al.(2017)Bastings, Titov, Aziz, Marcheggiani, and
  Sima'an}]{Bastings2017}
Joost Bastings, Ivan Titov, Wilker Aziz, Diego Marcheggiani, and Khalil
  Sima'an. 2017.
\newblock {Graph Convolutional Encoders for Syntax-aware Neural Machine
  Translation}.
\newblock In {\em Proceedings of EMNLP\/}. pages 1947--1957.

\bibitem[{Bojar et~al.(2017)Bojar, Graham, and Kamran}]{Bojar2017}
Ondřej Bojar, Yvette Graham, and Amir Kamran. 2017.
\newblock {Results of the WMT17 Metrics Shared Task}.
\newblock In {\em Proceedings of WMT\/}. volume~2, pages 293--301.

\bibitem[{Bruna et~al.(2014)Bruna, Zaremba, Szlam, and LeCun}]{Bruna2014}
Joan Bruna, Wojciech Zaremba, Arthur Szlam, and Yann LeCun. 2014.
\newblock {Spectral Networks and Locally Connected Networks on Graphs}.
\newblock In {\em Proceedings of ICLR\/}. page~14.

\bibitem[{Chen et~al.(2015)Chen, Li, Li, Lin, Wang, Wang, Xiao, Xu, Zhang, and
  Zhang}]{Chen2015}
Tianqi Chen, Mu~Li, Yutian Li, Min Lin, Naiyan Wang, Minjie Wang, Tianjun Xiao,
  Bing Xu, Chiyuan Zhang, and Zheng Zhang. 2015.
\newblock {MXNet: A Flexible and Efficient Machine Learning Library for
  Heterogeneous Distributed Systems}.
\newblock In {\em Proceedings of the Workshop on Machine Learning Systems\/}.
  pages 1--6.

\bibitem[{Chiang et~al.(2013)Chiang, Andreas, Bauer, Hermann, Jones, and
  Knight}]{Chiang2013}
David Chiang, Jacob Andreas, Daniel Bauer, Karl~Moritz Hermann, Bevan Jones,
  and Kevin Knight. 2013.
\newblock {Parsing Graphs with Hyperedge Replacement Grammars}.
\newblock In {\em Proceedings of ACL\/}. pages 924--932.

\bibitem[{Cho et~al.(2014)Cho, van Merrienboer, Gulcehre, Bahdanau, Bougares,
  Schwenk, and Bengio}]{Cho2014}
Kyunghyun Cho, Bart van Merrienboer, Caglar Gulcehre, Dzmitry Bahdanau, Fethi
  Bougares, Holger Schwenk, and Yoshua Bengio. 2014.
\newblock {Learning Phrase Representations using RNN Encoder-Decoder for
  Statistical Machine Translation}.
\newblock In {\em Proceedings of EMNLP\/}. pages 1724--1734.

\bibitem[{Drewes et~al.(1997)Drewes, Kreowski, and Habel}]{Drewes1997}
Frank Drewes, Hans~J{\"{o}}rg Kreowski, and Annegret Habel. 1997.
\newblock {Hyperedge Replacement Graph Grammars}.
\newblock {\em Handbook of Graph Grammars and Computing by Graph
  Transformation\/} .

\bibitem[{Duvenaud et~al.(2015)Duvenaud, Maclaurin, Aguilera-Iparraguirre,
  G{\'{o}}mez-Bombarelli, Hirzel, Aspuru-Guzik, and Adams}]{Duvenaud2015}
David Duvenaud, Dougal Maclaurin, Jorge Aguilera-Iparraguirre, Rafael
  G{\'{o}}mez-Bombarelli, Timothy Hirzel, Al{\'{a}}n Aspuru-Guzik, and Ryan~P
  Adams. 2015.
\newblock {Convolutional Networks on Graphs for Learning Molecular
  Fingerprints}.
\newblock In {\em Proceedings of NIPS\/}. pages 2215--2223.

\bibitem[{Eriguchi et~al.(2016)Eriguchi, Hashimoto, and
  Tsuruoka}]{Eriguchi2016}
Akiko Eriguchi, Kazuma Hashimoto, and Yoshimasa Tsuruoka. 2016.
\newblock {Tree-to-Sequence Attentional Neural Machine Translation}.
\newblock In {\em Proceedings of ACL\/}.

\bibitem[{Eriguchi et~al.(2017)Eriguchi, Tsuruoka, and Cho}]{Eriguchi2017}
Akiko Eriguchi, Yoshimasa Tsuruoka, and Kyunghyun Cho. 2017.
\newblock {Learning to Parse and Translate Improves Neural Machine
  Translation}.
\newblock In {\em Proceedings of ACL\/}.

\bibitem[{Flanigan et~al.(2016)Flanigan, Dyer, Smith, and
  Carbonell}]{Flanigan2016}
Jeffrey Flanigan, Chris Dyer, Noah~A. Smith, and Jaime Carbonell. 2016.
\newblock {Generation from Abstract Meaning Representation using Tree
  Transducers}.
\newblock In {\em Proceedings of NAACL\/}. pages 731--739.

\bibitem[{Flanigan et~al.(2014)Flanigan, Thomson, Carbonell, Dyer, and
  Smith}]{Flanigan2014}
Jeffrey Flanigan, Sam Thomson, Jaime Carbonell, Chris Dyer, and Noah~a Smith.
  2014.
\newblock {A Discriminative Graph-Based Parser for the Abstract Meaning
  Representation}.
\newblock In {\em Proceedings of ACL\/}.

\bibitem[{Galley et~al.(2004)Galley, Hopkins, Knight, and Marcu}]{Galley2004}
Michel Galley, Mark Hopkins, Kevin Knight, and Daniel Marcu. 2004.
\newblock {What's in a translation rule?}
\newblock In {\em Proceedings of NAACL\/}. pages 273--280.

\bibitem[{Gehring et~al.(2017)Gehring, Auli, Grangier, Yarats, and
  Dauphin}]{Gehring2017}
Jonas Gehring, Michael Auli, David Grangier, Denis Yarats, and Yann~N. Dauphin.
  2017.
\newblock {Convolutional Sequence to Sequence Learning}.
\newblock {\em arXiv preprint\/} .

\bibitem[{Glorot and Bengio(2010)}]{Glorot2010}
Xavier Glorot and Yoshua Bengio. 2010.
\newblock {Understanding the Difficulty of Training Deep Feedforward Neural
  Networks}.
\newblock In {\em Proceedings of AISTATS\/}. volume~9, pages 249--256.

\bibitem[{Gori et~al.(2005)Gori, Monfardini, and Scarselli}]{Gori2005}
Marco Gori, Gabriele Monfardini, and Franco Scarselli. 2005.
\newblock {A New Model for Learning in Graph Domains}.
\newblock In {\em Proceedings of IJCNN\/}. volume~2, pages 729--734.

\bibitem[{Graham et~al.(2014)Graham, Mathur, and Baldwin}]{Graham2014}
Yvette Graham, Nitika Mathur, and Timothy Baldwin. 2014.
\newblock {Randomized Significance Tests in Machine Translation}.
\newblock In {\em Proceedings of WMT\/}. pages 266--274.

\bibitem[{Gross and Yellen(2004)}]{Gross2004}
Jonathan Gross and Jay Yellen, editors. 2004.
\newblock {\em {Handbook of Graph Theory}\/}.
\newblock CRC Press.

\bibitem[{Hashimoto and Tsuruoka(2017)}]{Hashimoto2017}
Kazuma Hashimoto and Yoshimasa Tsuruoka. 2017.
\newblock {Neural Machine Translation with Source-Side Latent Graph Parsing}.
\newblock In {\em Proceedings of EMNLP\/}. pages 125--135.

\bibitem[{Hieber et~al.(2017)Hieber, Domhan, Denkowski, Vilar, Sokolov,
  Clifton, and Post}]{Hieber2017}
Felix Hieber, Tobias Domhan, Michael Denkowski, David Vilar, Artem Sokolov, Ann
  Clifton, and Matt Post. 2017.
\newblock {Sockeye: A Toolkit for Neural Machine Translation}.
\newblock {\em arXiv preprint\/} pages 1--18.

\bibitem[{Hochreiter and Schmidhuber(1997)}]{Hochreiter1997}
Sepp Hochreiter and J{\"{u}}rgen Schmidhuber. 1997.
\newblock {Long Short-Term Memory}.
\newblock {\em Neural Computation\/} 9(8):1735--1780.

\bibitem[{Jones et~al.(2012)Jones, Andreas, Bauer, Hermann, and
  Knight}]{Jones2012}
Bevan Jones, Jacob Andreas, Daniel Bauer, Karl~Moritz Hermann, and Kevin
  Knight. 2012.
\newblock {Semantics-Based Machine Translation with Hyperedge Replacement
  Grammars}.
\newblock In {\em Proceedings of COLING\/}. pages 1359--1376.

\bibitem[{Kearnes et~al.(2016)Kearnes, McCloskey, Berndl, Pande, and
  Riley}]{Kearnes2016}
Steven Kearnes, Kevin McCloskey, Marc Berndl, Vijay Pande, and Patrick Riley.
  2016.
\newblock {Molecular Graph Convolutions: Moving Beyond Fingerprints}.
\newblock {\em Journal of Computer-Aided Molecular Design\/} 30(8):595--608.

\bibitem[{Keskar et~al.(2017)Keskar, Mudigere, Nocedal, Smelyanskiy, and
  Tang}]{Keskar2017}
Nitish~Shirish Keskar, Dheevatsa Mudigere, Jorge Nocedal, Mikhail Smelyanskiy,
  and Ping Tak~Peter Tang. 2017.
\newblock {On Large-Batch Training for Deep Learning: Generalization Gap and
  Sharp Minima}.
\newblock In {\em Proceedings of ICLR\/}. pages 1--16.

\bibitem[{Kingma and Ba(2015)}]{Kingma2015}
Diederik~P. Kingma and Jimmy Ba. 2015.
\newblock {Adam: A Method for Stochastic Optimization}.
\newblock In {\em Proceedings of ICLR\/}. pages 1--15.

\bibitem[{Kipf and Welling(2017)}]{Kipf2017}
Thomas~N. Kipf and Max Welling. 2017.
\newblock {Semi-Supervised Classification with Graph Convolutional Networks}.
\newblock In {\em Proceedings of ICLR\/}.

\bibitem[{Koehn et~al.(2007)Koehn, Hoang, Birch, Callison-Burch, Federico,
  Bertoldi, Cowan, Shen, Moran, Zens, Dyer, Bojar, Constantin, and
  Herbst}]{Koehn2007}
Philipp Koehn, Hieu Hoang, Alexandra Birch, Chris Callison-Burch, Marcello
  Federico, Nicola Bertoldi, Brooke Cowan, Wade Shen, Christine Moran, Richard
  Zens, Chris Dyer, Ondrej Bojar, Alexandra Constantin, and Evan Herbst. 2007.
\newblock {Moses: Open source toolkit for statistical machine translation}.
\newblock In {\em Proceedings of ACL Demo Session\/}. pages 177--180.

\bibitem[{Konstas et~al.(2017)Konstas, Iyer, Yatskar, Choi, and
  Zettlemoyer}]{Konstas2017}
Ioannis Konstas, Srinivasan Iyer, Mark Yatskar, Yejin Choi, and Luke
  Zettlemoyer. 2017.
\newblock {Neural AMR: Sequence-to-Sequence Models for Parsing and Generation}.
\newblock In {\em Proceedings of ACL\/}. pages 146--157.

\bibitem[{Levi(1942)}]{Levi1942}
Friedrich~Wilhelm Levi. 1942.
\newblock {Finite Geometrical Systems}.

\bibitem[{Li et~al.(2016)Li, Tarlow, Brockschmidt, and Zemel}]{Li2016}
Yujia Li, Daniel Tarlow, Marc Brockschmidt, and Richard Zemel. 2016.
\newblock {Gated Graph Sequence Neural Networks}.
\newblock In {\em Proceedings of ICLR\/}. 1, pages 1--20.

\bibitem[{Liu et~al.(2006)Liu, Liu, and Lin}]{Liu2006}
Yang Liu, Qun Liu, and Shouxun Lin. 2006.
\newblock {Tree-to-string alignment template for statistical machine
  translation}.
\newblock In {\em Proceedings of the 21st International Conference on
  Computational Linguistics and the 44th annual meeting of the ACL - ACL
  '06\/}. pages 609--616.

\bibitem[{Luong et~al.(2015)Luong, Pham, and Manning}]{Luong2015}
Minh-Thang Luong, Hieu Pham, and Christopher~D. Manning. 2015.
\newblock {Effective Approaches to Attention-based Neural Machine Translation}.
\newblock In {\em Proceedings of EMNLP\/}. pages 1412--1421.

\bibitem[{Marcheggiani and Titov(2017)}]{Marcheggiani2017}
Diego Marcheggiani and Ivan Titov. 2017.
\newblock {Encoding Sentences with Graph Convolutional Networks for Semantic
  Role Labeling}.
\newblock In {\em Proceedings of EMNLP\/}.

\bibitem[{Och and Ney(2002)}]{Och2002a}
Franz~Josef Och and Hermann Ney. 2002.
\newblock \href{https://doi.org/10.3115/1073083.1073133}{{Discriminative
  training and maximum entropy models for statistical machine translation}}.
\newblock In {\em Proceedings of the 40th Annual Meeting on Association for
  Computational Linguistics - ACL '02\/}. page 295.
\newblock
  \href{https://doi.org/10.3115/1073083.1073133}{https://doi.org/10.3115/1073083.1073133}.

\bibitem[{Papineni et~al.(2001)Papineni, Roukos, Ward, and Zhu}]{Papineni2001}
Kishore Papineni, Salim Roukos, Todd Ward, and Wei-Jing Zhu. 2001.
\newblock {Bleu: a method for automatic evaluation of machine translation}.
\newblock In {\em Proceedings of ACL\/}. pages 311--318.

\bibitem[{Peng et~al.(2015)Peng, Song, and Gildea}]{Peng2015}
Xiaochang Peng, Linfeng Song, and Daniel Gildea. 2015.
\newblock {A Synchronous Hyperedge Replacement Grammar based approach for AMR
  parsing}.
\newblock In {\em Proceedings of CoNLL\/}. pages 32--41.

\bibitem[{Popovi{\'{c}}(2017)}]{Popovic2017}
Maja Popovi{\'{c}}. 2017.
\newblock {chrF ++: words helping character n-grams}.
\newblock In {\em Proceedings of WMT\/}. pages 612--618.

\bibitem[{Pourdamghani et~al.(2014)Pourdamghani, Gao, Hermjakob, and
  Knight}]{Pourdamghani2014}
Nima Pourdamghani, Yang Gao, Ulf Hermjakob, and Kevin Knight. 2014.
\newblock {Aligning English Strings with Abstract Meaning Representation
  Graphs}.
\newblock In {\em Proceedings of EMNLP\/}. pages 425--429.

\bibitem[{Pourdamghani et~al.(2016)Pourdamghani, Knight, and
  Hermjakob}]{Pourdamghani2016}
Nima Pourdamghani, Kevin Knight, and Ulf Hermjakob. 2016.
\newblock {Generating English from Abstract Meaning Representations}.
\newblock In {\em Proceedings of INLG\/}. volume~0, pages 21--25.

\bibitem[{Reimers and Gurevych(2017)}]{Reimers2017}
Nils Reimers and Iryna Gurevych. 2017.
\newblock {Reporting Score Distributions Makes a Difference: Performance Study
  of LSTM-networks for Sequence Tagging}.
\newblock In {\em Proceedings of EMNLP\/}. pages 338--348.

\bibitem[{Scarselli et~al.(2009)Scarselli, Gori, Tsoi, and
  Monfardini}]{Scarselli2009}
Franco Scarselli, Marco Gori, Ah~Ching Tsoi, and Gabriele Monfardini. 2009.
\newblock {The Graph Neural Network Model}.
\newblock {\em IEEE Transactions on Neural Networks\/} 20(1):61--80.

\bibitem[{Schlichtkrull et~al.(2017)Schlichtkrull, Kipf, Bloem, van~den Berg,
  Titov, and Welling}]{Schlichtkrull2017}
Michael Schlichtkrull, Thomas~N. Kipf, Peter Bloem, Rianne van~den Berg, Ivan
  Titov, and Max Welling. 2017.
\newblock {Modeling Relational Data with Graph Convolutional Networks} pages
  1--12.

\bibitem[{Sennrich et~al.(2016)Sennrich, Haddow, and Birch}]{Sennrich2016}
Rico Sennrich, Barry Haddow, and Alexandra Birch. 2016.
\newblock {Neural Machine Translation of Rare Words with Subword Units}.
\newblock In {\em Proceedings of ACL\/}. pages 1715--1725.

\bibitem[{Song et~al.(2017)Song, Peng, Zhang, Wang, and Gildea}]{Song2017}
Linfeng Song, Xiaochang Peng, Yue Zhang, Zhiguo Wang, and Daniel Gildea. 2017.
\newblock {AMR-to-text Generation with Synchronous Node Replacement Grammar}.
\newblock In {\em Proceedings of ACL\/}. pages 7--13.

\bibitem[{Srivastava et~al.(2014)Srivastava, Hinton, Krizhevsky, Sutskever, and
  Salakhutdinov}]{Srivastava2014}
Nitish Srivastava, Geoffrey Hinton, Alex Krizhevsky, Ilya Sutskever, and Ruslan
  Salakhutdinov. 2014.
\newblock {Dropout: A Simple Way to Prevent Neural Networks from Overfitting}.
\newblock {\em Journal of Machine Learning Research\/} 15:1929--1958.

\bibitem[{Vaswani et~al.(2017)Vaswani, Shazeer, Parmar, Uszkoreit, Jones,
  Gomez, Kaiser, and Polosukhin}]{Vaswani2017}
Ashish Vaswani, Noam Shazeer, Niki Parmar, Jakob Uszkoreit, Llion Jones,
  Aidan~N. Gomez, Lukasz Kaiser, and Illia Polosukhin. 2017.
\newblock {Attention Is All You Need}.
\newblock In {\em Proceedings of NIPS\/}.

\bibitem[{Wu(1997)}]{Wu1997}
Dekai Wu. 1997.
\newblock {Stochastic inversion transduction grammars and bilingual parsing of
  parallel corpora}.
\newblock {\em Computational Linguistics\/} 23(3):377--403.

\bibitem[{Yamada and Knight(2001)}]{Yamada2001}
Kenji Yamada and Kevin Knight. 2001.
\newblock {A Syntax-based Statistical Translation Model}.
\newblock In {\em Proceedings of ACL\/}. pages 523--530.

\end{thebibliography}
\bibliographystyle{acl_natbib}

\appendix

\section{Simplification and Anonymisation for AMR graphs}
\label{sec:simpl-anonym-amr}

The procedure for graph simplification and anonymisation is similar to what is done by \cite{Konstas2017}. The main difference is that we use the alignments provided by the original LDC version of the AMR corpus, while they use a combination of the JAMR aligner \cite{Flanigan2014} and the unsupervised aligner of \cite{Pourdamghani2014}. This preprocessing is done before transforming the graph into its bipartite equivalent.
\begin{description}
\item[Simplification:] we remove sense information from the concepts. For instance, {\tt believe-01} becomes {\tt believe}. We also remove any subgraphs related to wikification (the ones starting the predicate {\tt :wiki}).
\item[Entity anonymisation:] all subgraphs starting with the predicate {\tt :name} are anonymised. The predicate contains one of the AMR entity names as the source node, such as {\tt country} or {\tt *-quantity}. These are replaced by a single anonymised node, containing the original entity concept plus an index. At training time, we use the alignment information to find the corresponding entity in the surface form and replace all aligned tokens with the same concept name. At test time, we extract a map that maps the anonymised entity to all concept names in the subgraph, in depth-first order. After prediction, if there is an anonymised token in the surface form it is replaced using that map (as long as the predicted token is present in the map).
\item[Enitity clustering:] entities are also clustered into four coarse-grained types in both graph and surface form. For instance {\tt country\_0} becomes {\tt loc\_0}. We use the list obtained from the open source implementation of \cite{Konstas2017} for that.
\item[Date anonymisation:] if there is a {\tt date-entity} concept, all underlying concepts are also anonymised, using separate tokens for day, month and year. At training time, the surface form is also anonymised following the alignment, but we additionally split days and months into {\tt day\_name, day\_number, month\_name} and {\tt month\_number}. At test time, we render the day/month according to the predicted anonymised token in the surface form and the recorded map.
\end{description}

\section{Model Hyperparameters}
\label{sec:model-hyperp}

Our implementation is based on the Sockeye toolkit for Neural Machine Translation \cite{Hieber2017}. Besides the specific hyperparameter values mentioned in the paper, all other hyperparameters are set to the default values in Sockeye. We detail them here for completeness:

\subsection{Vocabulary}
\begin{itemize}
\item For AMR, we set the minimum frequency to 2 in both source nodes and target surface tokens. For NMT, we also use 2 as the minimum frequency in the source but use 1 in the target since we use BPE tokens.
\end{itemize}

\subsection{Model}
\label{sec:model}
\begin{itemize}
\item The baseline encoder use a BiLSTM followed by a unidirectional LSTM. The decoder in all models use a 2-layer LSTM.
\item The attention module uses a bilinear scoring function ({\em general} as in \cite{Luong2015}).
\item The max sequence length during training is 200 for AMR. In NMT, we use 100 for the \stos baselines and 200 for the \gtos models. This is because we do not use dependency trees in the baseline so it naturally has half of the tokens.
\item All dimensionalities are fixed to 512, which is similar to what is used by \cite{Konstas2017}. The only exceptions are the GGNN hidden state dimensionality in the \gtos models. We use 576 for the \gtos models used in the AMR experiments and 448 for the \gtosplus models used in the NMT experiments. As pointed out in the main paper, we change GGNN dimensionalities in order to have a similar parameter budget compared to the \stos baselines.
\end{itemize}

\subsection{Training}
\label{sec:training}

These options apply for both \stos baselines and \gtos \ \gtosplus models.

\begin{itemize}
\item We use 16 as the batch size. This is a lower number than most previous work we compare with: we choose this because we obtained better results in the AMR dev set. This is in line with recent evidence showing that smaller batch sizes lead to better generalisation performance \cite{Keskar2017}. The drawback is that smaller batches makes training time slower. However, this was not a problem in our experiments due to the medium size of the datasets.
\item Bucketing is used to speed up training: we use 10 as the bucket size.
\item Models are trained using cross-entropy as the loss.
\item We save parameter checkpoints at every full epoch on the training set.
\item We use early stopping by perplexity on the dev set with patience 8 (training stops if dev perplexity do not improve for 8 checkpoints).
\item A maximum of 30 epochs/checkpoints is used. All our models stopped training before reaching this limit.
\item We use 0.5 dropout on the input embeddings, before they are fed to the encoder.
\item Weigths are initalised using Xavier initialisation \cite{Glorot2010}, with except of forget biases in the LSTMs which are initialised by 0.
\item We use Adam \cite{Kingma2015} as the optimiser with 0.0003 as the initial learning rate.
\item Learning rate is halved every time dev perplexity does not improve for 3 epochs/checkpoints.
\item Gradient clipping is set to 1.0.
\end{itemize}

\subsection{Decoding}
\label{sec:decoding}

\begin{itemize}
\item We use beam search to decode, using 5 as the beam size.
\item Ensembles are created by averaging log probabilities at every step ({\em linear} in Sockeye). Attention scores are averaged over the 5 models at every step.
\item For AMR only, we replace {\tt <unk>} tokens in the prediction with the node with the highest attention score at that step, in the same way described by \cite{Konstas2017}. This is done before deanonymisation.
\end{itemize}

\end{document}